\documentclass[sigconf]{acmart}

\usepackage{booktabs} 

\usepackage{subcaption, mwe}

\usepackage{booktabs}

\usepackage{todonotes}

\usepackage{here}
\usepackage{graphicx}
\usepackage{caption}
\usepackage[ruled,noend]{algorithm2e}
\usepackage{amsmath,amssymb,amsfonts,amsbsy,amsfonts,latexsym}
\usepackage{multirow}
\usepackage{makecell}

\usepackage{subcaption}
\usepackage[labelfont=bf,textfont=it,belowskip=0pt,aboveskip=5pt,tableposition=top]{caption}
\usepackage{xcolor}
\usepackage{colortbl}

\definecolor{colorA}{RGB}{189,201,225}
\definecolor{colorB}{RGB}{103,169,207}
\definecolor{colorC}{RGB}{ 28,144,153}
\definecolor{colorD}{RGB}{  1,108, 89}

\newcolumntype{R}{>{\columncolor{gray!40}}r}
\newcolumntype{L}{>{\columncolor{gray!40}}l}
\newcolumntype{C}{>{\columncolor{gray!40}}c}

\usepackage{xparse}

\graphicspath{{figs}}
\DeclareGraphicsExtensions{.pdf,.png}
\usepackage{pgfplots, pgfplotstable}

\usepackage{mathtools}
\DeclarePairedDelimiter{\ceil}{\lceil}{\rceil}
\DeclarePairedDelimiter{\floor}{\lfloor}{\rfloor}

\NewDocumentCommand{\var}{O{s} m O{}}{%
  \ensuremath{#1_{#2}^{#3}}
}

\renewcommand{\d}[1]{\mathop{}\!\mathrm{d}#1}

\definecolor{light-gray}{gray}{0.80}

\renewcommand\paragraph{\subsubsection*}

\newcommand\eref{Eq. \ref}
\newcommand\fref{Fig. \ref}

\SetKwInOut{Parameter}{parameter}

\newcommand\J{\mathcal{J}}

\begin{document}

\title{Integrated Model, Batch, and Domain Parallelism in Training Neural Networks}

\author{Amir Gholami}
\affiliation{%
  \institution{EECS Department, UC Berkeley}
}
\email{amirgh@eecs.berkeley.edu}

\author{Ariful Azad}
\affiliation{%
  \institution{CRD, Lawrence Berkeley Lab}
}
\email{azad@lbl.gov}

\author{Peter Jin}
\affiliation{%
  \institution{EECS Department, UC Berkeley}
}
\email{phj@eecs.berkeley.edu}

\author{Kurt Keutzer}
\affiliation{%
  \institution{EECS Department, UC Berkeley}
}
\email{keutzer@eecs.berkeley.edu}

\author{Ayd\i{}n Bulu\c{c}}
\affiliation{%
  \institution{CRD, Lawrence Berkeley Lab} 
}
\email{abuluc@lbl.gov}

\acmConference{\textcopyright [Gholami et al.] [2018] This is the author's version of the work. It is posted here for your personal use. Not for redistribution. The definitive version was published in SPAA '18: 30th ACM Symposium on Parallelism in Algorithms and Architectures}
\acmPrice{}
\acmDOI{10.1145/3210377.3210394}
\acmISBN{}
\setcopyright{none}
\settopmatter{printacmref=false}
\pagestyle{plain} 

\begin{abstract}
  We propose a new integrated method of exploiting model, batch and domain
  parallelism for the training of deep neural networks (DNNs) on large
  distributed-memory computers using minibatch stochastic gradient descent
  (SGD). Our goal is to find an efficient parallelization strategy for a fixed
  batch size using $P$ processes.  Our method is inspired by the
  communication-avoiding algorithms in numerical linear algebra. We see $P$
  processes as logically divided into a $P_r \times P_c$ grid where the $P_r$
  dimension is implicitly responsible for model/domain parallelism and the $P_c$
  dimension is implicitly responsible for batch parallelism. In practice, the
  integrated matrix-based parallel algorithm encapsulates these types of
  parallelism automatically. We analyze the communication complexity and
  analytically demonstrate that the lowest communication costs are often
  achieved neither with pure model nor with pure data parallelism.
  We also show how the domain parallel approach can help in extending 
  the theoretical scaling limit of the typical batch parallel method.
	
\end{abstract}



\maketitle

\section{Introduction and Background}


Neural Networks (NNs) have proved to be very effective in diverse applications ranging
from semantic segmentation~\cite{long2015fully,squeezeseg} and detection~\cite{ren2015faster,wu2016squeezedet} to
medical image segmentation~\cite{havaei2017brain,brats}.
In most cases the hardware limits have been reached for most of the kernels, and the next milestone
is in distributed computing. This is becoming increasingly important with renewed attention to super resolution machine learning~\cite{kim2016accurate},
as well as significant increase in the training dataset in cases such as autonomous driving. Effective use of these datasets in a reasonable time is
not possible without a scalable parallel method.

Given $N$ empirical samples, the DNN training procedure seeks to find the model
parameters, $w$, such that the forward pass on sample inputs would produce outputs that are \textit{similar} to ground truth outputs and that it generalizes
well for unseen test samples. 
The weights
are initialized randomly and SGD algorithm updates them iteratively as:  $w^{n+1} = w^{n} - \eta \nabla f_i$, where $i$ is an index chosen randomly (with replacement) from $[1,N]$, $\eta$ is the learning rate, and $f$ is the loss function. In practice, one can use a mini-batch SGD by drawing a set of indices $i\in \textrm{Batch}$ at each iteration, chosen randomly
from $[1,N]$ and update the parameters as follows:
\begin{align}
  w^{n+1} = w^{n} - \eta \frac{1}{B}\sum_{i\in \textrm{Batch}}\nabla f_i,
\label{e:sgd}
\end{align}

\noindent where $B$ is the mini-batch size. 
This whole SGD-based training requires a ``forward pass'' where the network's output and the
corresponding loss functional is computed given the current model parameters,
and a ``backward pass'' (commonly referred to as {\em backpropagation} or simply {\em backprop}) where the gradient of the loss is computed with respect
to the model parameters, $w$.

The forward phase of DNN training is a sequential combination of
affine transformation $Y_i = W_i X_i$, followed by
nonlinear transforms $X_{i+1} = f(Y_i)$ . 
Each column of $X_i \in \mathbb{R}^{d_{i-1} \times B }$ holds input activations for one sample and similarly each
column of $Y_i \in {\mathbb{R}^{d_i \times B}}$ holds output activations for one sample. Notice that $X_{i+1}$ and $Y_i$ have the same shape. 
The matrix $W_i \in \mathbb{R}^{d_i \times d_{i-1}}$ holds the weights of the neural network between the $i$th and $(i-1)$th layer. 
The number of neurons in the $i$th DNN layer is denoted by $d_i$.

Forward phase is followed by
backpropagation that can also be written in matrix form as $\Delta_{X_i} = W_i^T
\Delta_{Y_i}$. Here, $\Delta_{X_i}$ and $\Delta_{Y_i}$ are the gradients of the loss
function, with respect to input and output activations, respectively.  Finally,
the gradient of the loss function with respect to model weights is calculated
using $\Delta_{W_i} = \Delta_{Y_i} X_i^T$. Consequently, DNN training requires 3 \textit{matrix multiplications}, including gradient computations.\footnote{Note that our approach
does not require each individual convolution to be
computed using matrix multiplication, but we view it as
this way for simplicity and connection to high performance
computing literature.}
The derivations of the forward pass and the backpropagation are shown in detail in Sections~\ref{forwardpass} and \ref{backprop}, respectively.

A single pass over the whole data (also called an {\em epoch}) requires $N/B$
iterations. It takes many iterations until the training error is sufficiently
small. Consequently, DNN training is computationally expensive. To accelerate
training, one can change the training algorithm with an aim to reduce the
number of epochs, or make each epoch run faster through distributed training.
We are focusing on the latter.

\begin{figure}[t]
    \centering
    \includegraphics[width=0.46  \textwidth]{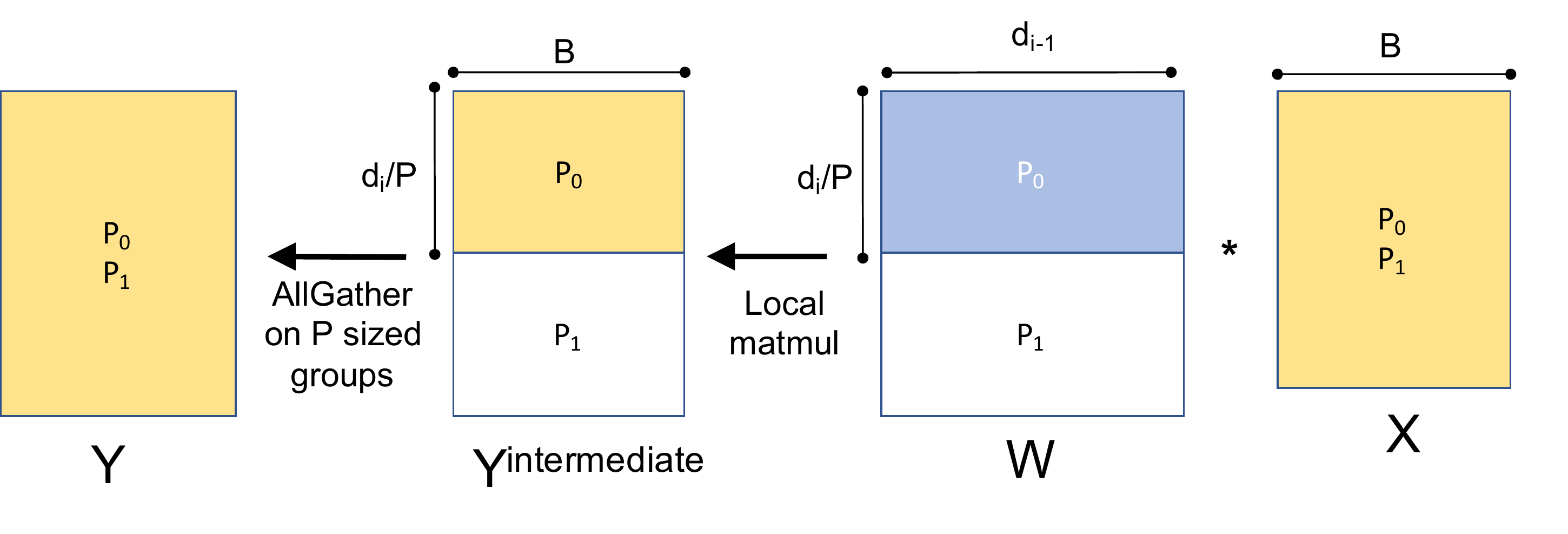}\\
    \includegraphics[width=0.39 \textwidth] {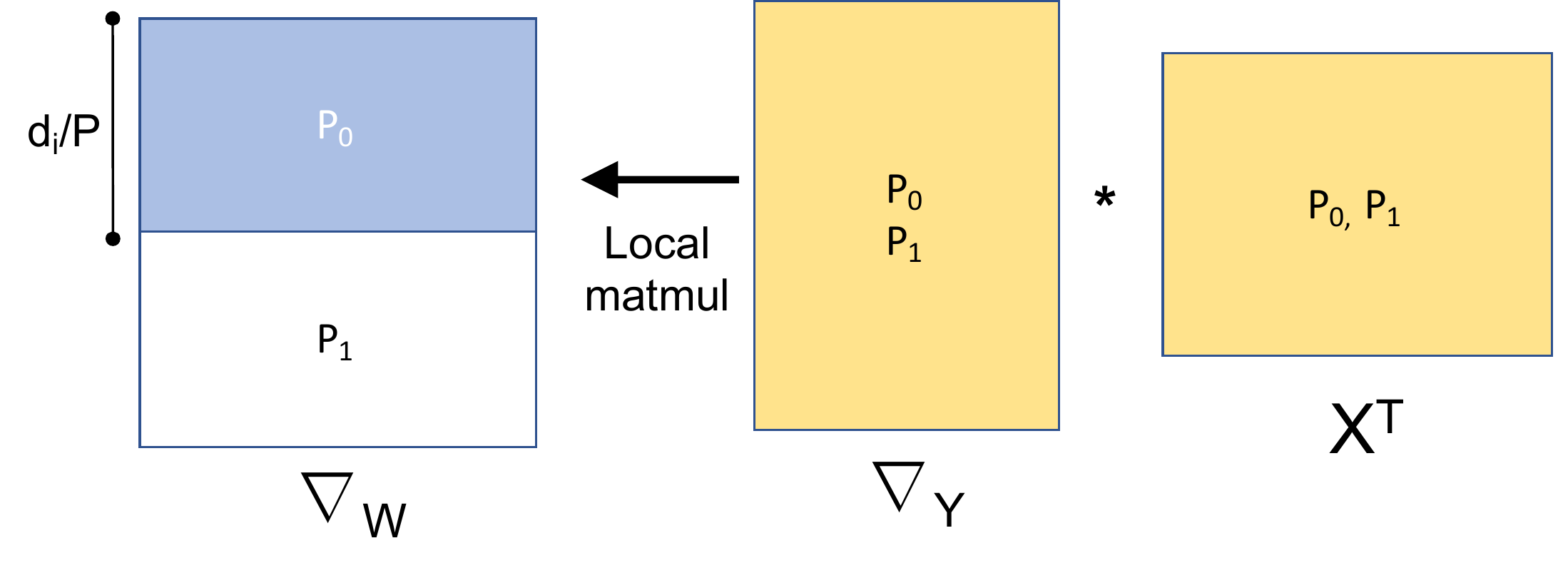}\qquad\qquad\qquad\\
    \includegraphics[width=0.46  \textwidth]{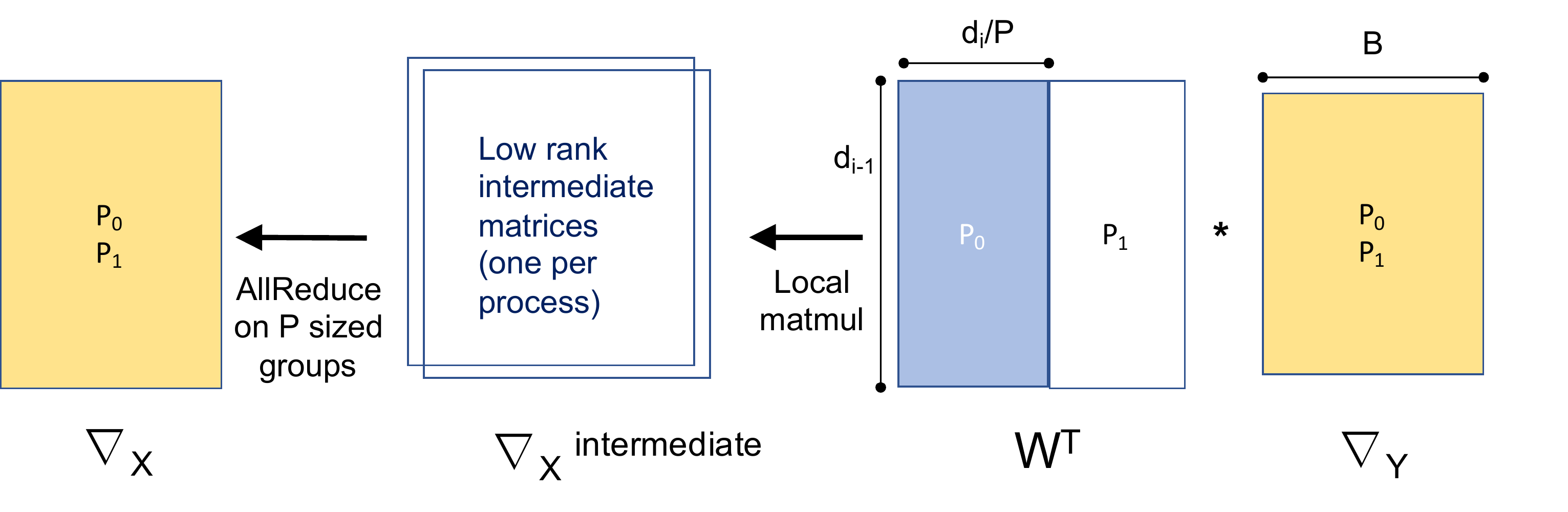}
    \captionof{figure}{ Illustration of matrix multiplications for the pure model parallel training using $P=2$ (top: forward pass, middle/bottom: weight gradient
  computation).}
    \label{fig:model}
\end{figure}

Two well-known techniques for distributed SGD based DNN training are {\em model parallelism} and {\em data parallelism}. 
In simplest terms, model parallelism is the partitioning of the weights of the neural network to processes.
Data parallelism corresponds to partitioning of the input data to processes. The existing literature merely considers 
data parallelism to be the assignment
of groups of whole data points, such as images, to individual processes. However, one can instead assign fractions of data
points to processes as well. For example, training a convolutional neural network (CNN) on two processes with domain parallelism can assign 
all the top halves of the images to the first processor and all
the bottom halves of the images to the second processor~\cite{jin2018spatially}. Consequently, there are two subtypes of data parallelism:
{\em batch parallelism}, which is the commonly studied option in literature, is the assignment of groups of data points 
in whole to processes and {\em domain parallelism} is the subdivision of individual data points to processes.

This paper presents a new method for integrated model, batch, and domain parallelism.
There are existing approaches that exploit both model and batch parallelism but
they often only provide ad-hoc solutions to hard engineering constraints such
as the model no longer fitting into a single GPU or the mini batch sizes hitting
a convergence limit. Our method, by contrast, is amenable to precise
communication analysis and covers the whole spectrum between pure data
parallelism (which includes batch parallelism as a special case) and pure model parallelism. 
It often finds favorable performance regimes that are better than pure batch parallelism and 
pure model parallelism, even in the absence of hard engineering constraints. 

\paragraph{Limitations} We find it useful and necessary to describe the
limitations in our analysis.  For the communication complexity we assume that
all the compute nodes are connected and thus do not consider the topology of
the interconnect, and we also do not consider network conflicts in our
model~\cite{chan2007collective}. However, the effects of this can be
approximated by adjusting the latency and bandwidth terms accordingly, as
a detailed analysis will become network specific.  

While the presented simulated results
are based on AlexNet, the mathematical analysis we present for the integrated framework is generally applicable to any neural network. 
For instance, cases with Recurrent Neural Networks mainly consist of fully connected layers and
our analysis naturally extends to those cases.
Moreover, we empirically measure the computation time. A more detailed
analysis of the computation time would require a hardware specific execution
model which is outside the scope of this work. Finally, we present simulation
results based on the complexity analysis. Those simulation results assume idealized network behavior (i.e. perfect utilization of bandwidth, no additional 
software overheads, and perfect overlap of communication and computation when considered), and hence provide an upper bound on achievable performance.


\section{Parallelism in DNN Training}

Deep Neural Networks are typically trained using first-order methods, i.e. those that rely on 
first order derivatives. SGD is the canonical example of first-order methods used in DNN training. 
Regardless of the specific approach, all methods calculate activations using forward propagation and
calculate derivatives using backprop. Consequently, our results generalize to other first-order methods
even though we will describe it using SGD for simplicity.

The SGD iterations have a sequential dependency. One possibility to break this
barrier for parallel training is the family of asynchronous SGD
methods~\cite{rogers1998using,jin2016scale,recht2011hogwild,chilimbi2014project,dean2012large,zhang2015deep}.
Here, this dependency is broken and each process is allowed to use
\textit{stale} parameters and update either its weights or that of a parameter
server. However, these approaches often do not converge to the same performance as in
the synchronous SGD cases.
Here, we focus only on the latter which obeys the
sequential consistency of the original algorithm. However, the framework that
we present can be used to accelerate asynchronous methods as well.

In terms of terminology, we use the word ``process" to refer to the program
running on a compute node. It is often the case that a compute node has many
processing elements (or cores); thus one can map multiple processes, each with
its own local private memory, to a compute node. The exact nature of process to
compute node mapping is immaterial to our analysis. 

\begin{figure}[t]
    \centering
    \includegraphics[width=0.36  \textwidth]{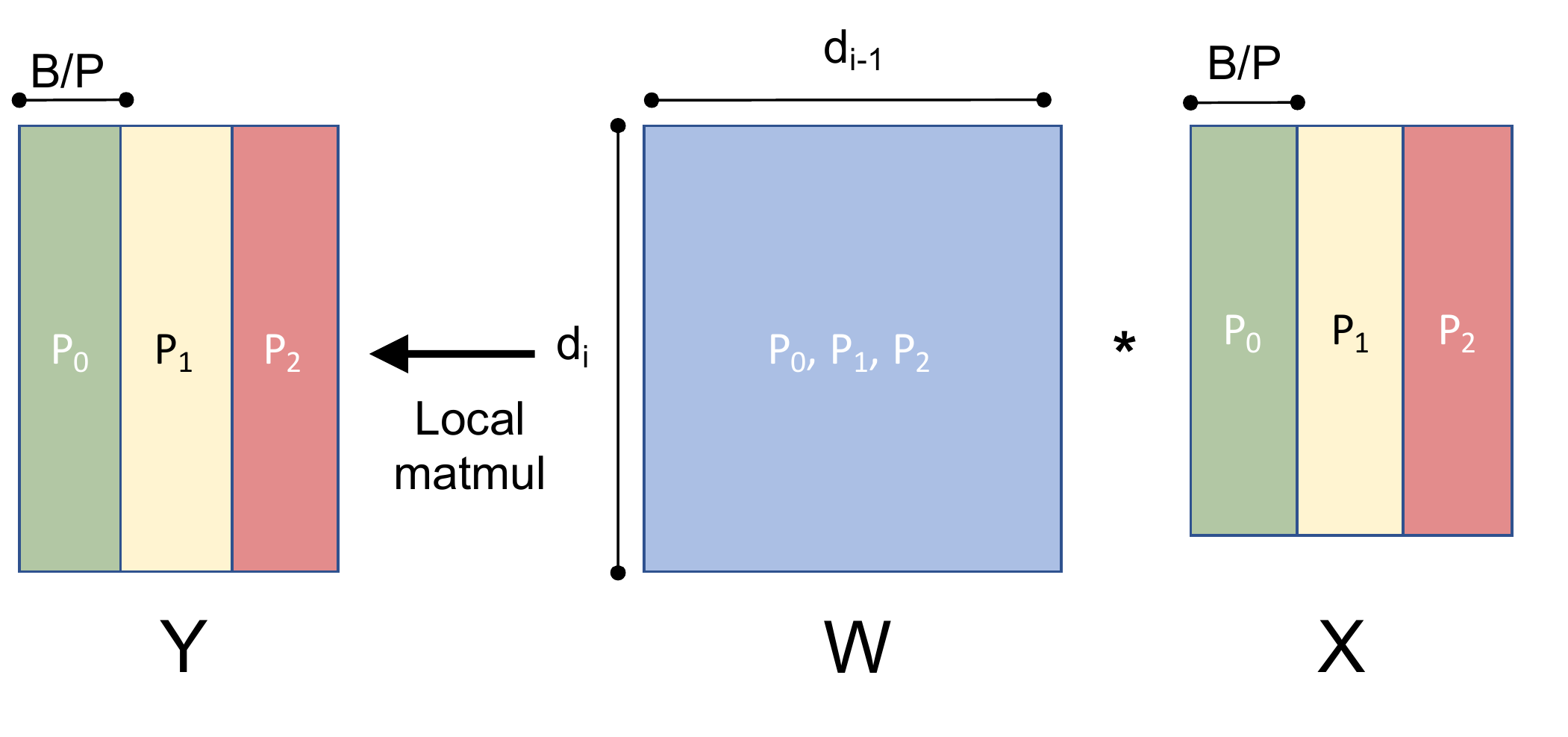}\\
    \includegraphics[width=0.47  \textwidth]{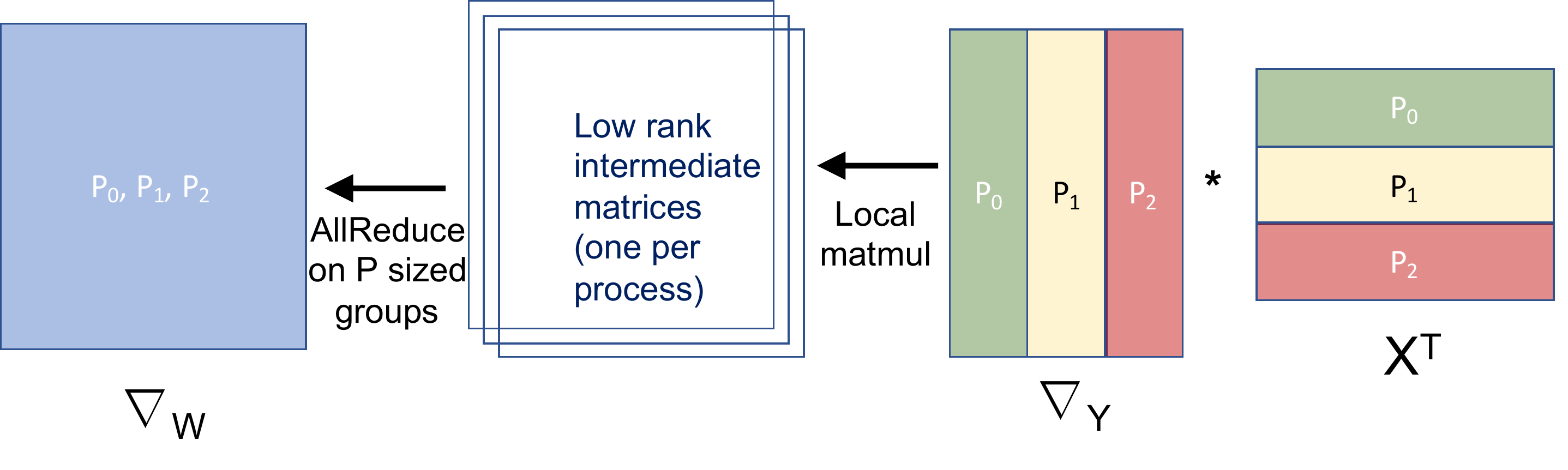}\\
    \includegraphics[width=0.36  \textwidth]{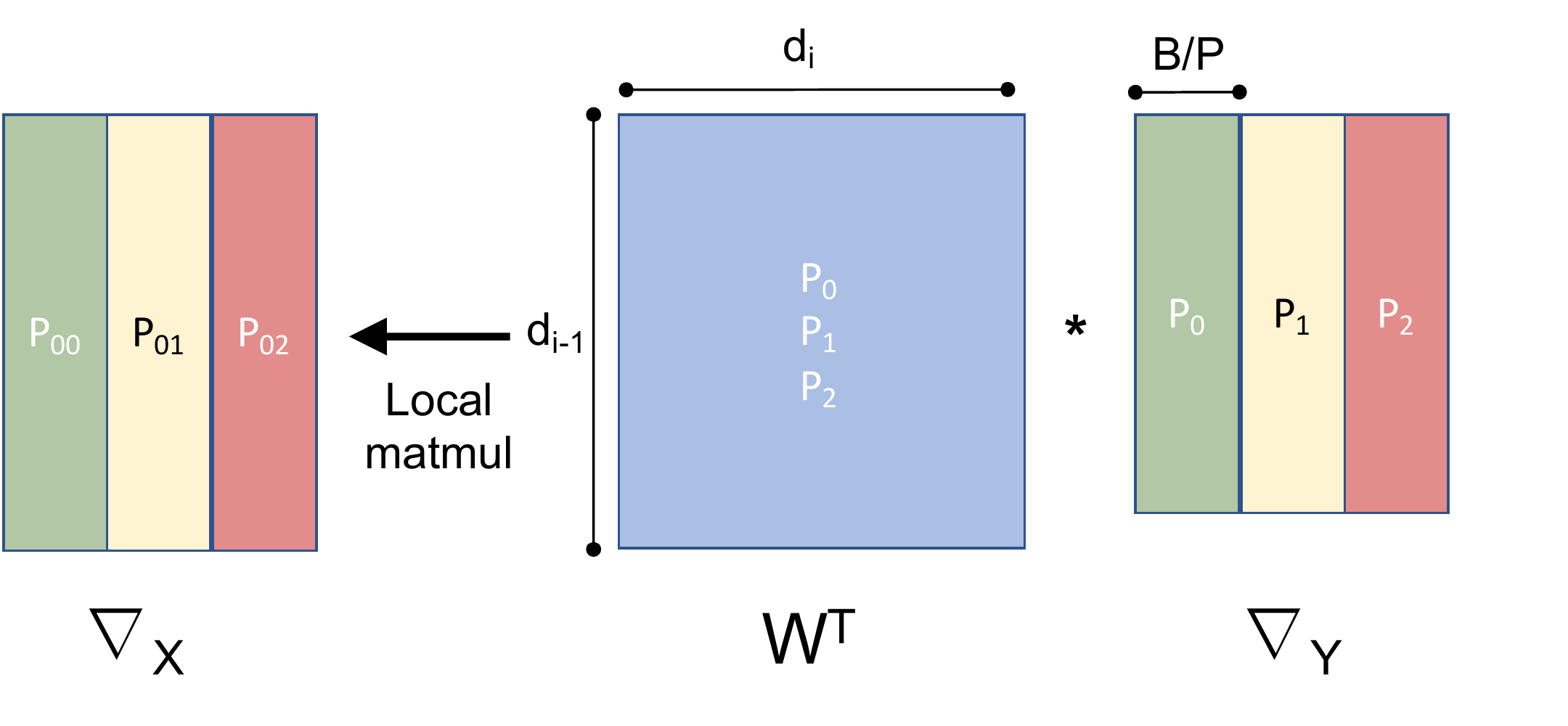}
    \captionof{figure}{ Illustration of matrix multiplications for the pure batch parallel training using $P=3$ (top: forward pass, middle/bottom: weight gradient
  computation).}
    \label{fig:data}  
\end{figure}




\subsection{Layers of Deep Neural Networks}
Deep Neural Networks are composed of many layers. Typically each layer is either a convolutional layer, a fully connected layer, activation layer, or a dropout layer.
A convolutional layer is composed of a number of filters (also called kernels), applied in a sliding window fashion with a stride length $s$ over the whole input sample.
The application of each filter in a convolutional layer results in a distinct {\em channel} in the output layer. Hence, we will use $X^i_C$ to denote
the number of channels in the $i$th layer. The number of input channels in the first layer is equal to the number of channels in the input data (usually three channels for RGB). 
A convolutional filter in the $i$th layer takes a tensor input $k_h^i \times k_w^i \times X^i_C$ and creates a single scalar value (Here $k_h^i,\ k_w^i$ are the kernel
convolution kernel's size). There are
$Y^i_C$ such different filters in the $i$th convolutional layer. Consequently an input of dimensions $X^i_H,\ X^i_W,\ X^i_C$ is transformed into
an output of dimensions $Y^i_W,\ Y^i_H,\ Y^i_C$ where
$$ Y^i_W = \ceil*{\frac{X^i_W-k_w}{s}}, Y^i_H = \ceil*{\frac{X^i_H-k_h}{s}}. $$
With proper padding, it simplifies to $Y^i_W = \ceil*{X^i_W/s}$ and $Y^i_H = \ceil*{X^i_H/s}$.

The number of distinct parameters between two convolutional layers is equal to the number of nonzeros in $W$ if they are represented compactly without redundancy.
Hence, 
\begin{equation}
\begin{split}
	\lvert W_i \rvert &= (k_h k_w X^i_C) Y^i_C, \\
	d_{i-1} &= X^i_H X^i_W X^i_C,\\
	d_i &= Y^i_H Y^i_W Y^i_C = \ceil*{X^i_W/s} \ceil*{X^i_H/s} Y^i_C. 
	\end{split}
		\label{eq:parameters}
\end{equation}

The number of parameters between two fully-connected layers, or between a convolutional layer and a fully-connected layer is simply $\lvert W_i \rvert = d_i d_{i-1}$.
Dropout is sometimes applied to fully-connected layers and has the effect of pruning 
a certain percentage of both the input and output activations.  

\subsection{Communication Cost Analysis of Pure Batch, Pure Model, and Pure Domain-Parallel Approaches}

Two possibilities for parallel computations in synchronous SGD is model and
data parallel. The latter can be subdivided into batch parallelism and domain parallelism as explained in the previous section.

{\bf Communication costs of pure model parallelism.} 
In the model parallel case, the computation of the loss
in the forward pass can be computed by distributing the model
parameters $W$ as shown in \fref{fig:model}.

\begin{figure*}[tb]
    \centering
    \includegraphics[width=0.75  \textwidth]{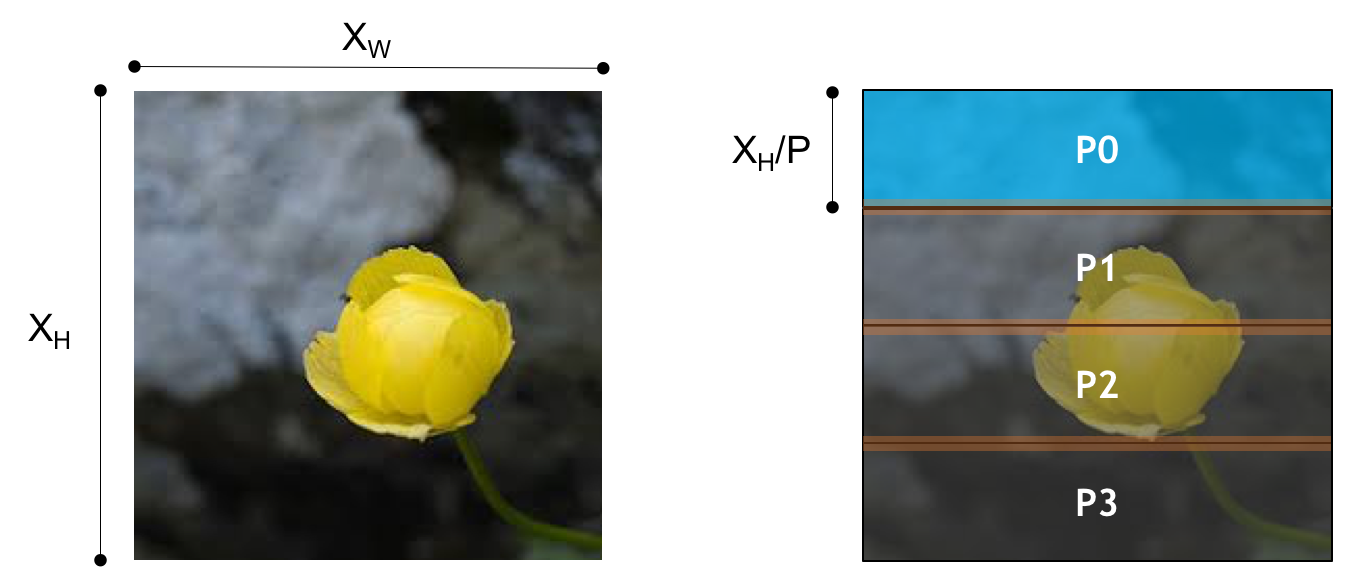}
    \captionof{figure}{Illustration of domain parallel approach for $P=4$. For NCHW format, it is best to
    distribute along the height to avoid non-contiguous memory accesses. NCHW format corresponds to the data layout in the memory,
    where the data runs fastest in width, height, channel size, and then across batch size.
  }
    \label{fig:domain}  
\end{figure*}

Consider a convolutional layer without loss of generality: each process performs a
subset of the convolutions on the input activations and computes a subset of
the output activations. For instance, assume one of the layers consists of
$Y_C$ $k_h\times k_w\times X_C$ convolutions, where $k_h,\ k_w$ is the
size of each convolution filter and $X_C,\ Y_C$ are the sizes of input and output channels. 
In the model parallel case, the kernels are distributed so that each process gets
$Y_C/P$ filters 
and computes the corresponding $Y_C/P$ channels of the output activation.  As
computations of the other layers would require access to all of the previous
activations, one needs to perform
an all-gather operation {\em per layer}. Backpropagation also requires an
all-reduce communication during $\Delta_X$ calculation (details are discussed
in appendix). This yields the following communication complexity for the model
parallel case:


\begin{equation}
\begin{split}
  T_{\mathit{comm}}(\mathit{model}) &= \sum_{i=1}^{L}\left(\alpha \ceil{\log(P)} + \beta B \frac{P-1}{P} d_i \right) \\
  &+ 2\sum_{i=2}^{L}\left(\alpha \ceil{\log(P)} + \beta B \frac{P-1}{P} d_{i-1}\right),
  \end{split}
\end{equation}
\noindent where $P$ is the number of processes, $L$ is the number of DNN
layers, $\alpha$ is the network latency, and $\beta$ is the inverse bandwidth.
The first sum considers the cost for all-gather required after every layer, and
the second sum considers the all-reduce cost for backpropagating activation gradients.
Note that the second sum starts from $i=2$ as we do not need to backpropagate the
gradient beyond the first layer.
This analysis assumes the use of Bruck's algorithm for all-gather and ring
algorithm for all-reduce~\cite{thakur2005optimization}. We note that the
complexity depends on the mini-batch size. The model parallel approach was
partially used in AlexNet~\cite{krizhevsky2012imagenet}, where the model was
split into two GPUs. The original GoogLeNet work also exploited a certain
amount of model parallelism~\cite{szegedy2015going}. Distributed DNN training
engines that rely solely on model parallelism also exist~\cite{coates2013deep},
especially for low-latency high-bandwidth systems.  

The other possibility for distributing the SGD computation is data parallelism.
This can be performed either by distributing the data over the batch size, or partition each
individual image. We refer to the latter as {\em domain parallelism}, which will be discussed further below.

{\bf Communication costs of pure batch parallelism.} 
For the batch parallel case, the reduction for the gradient computation over the mini-batch
sum~\eqref{e:sgd} can be computed independently by each process.  This approach
is known as \textit{batch parallel} method, where each process computes a
partial sum, followed by an all-reduce to compute the mini-batch gradient.
This communication cost is due to the reduction that is needed to form
$\Delta_W = \Delta_Y X^T $ product.  The communication complexity for the batch
parallel approach using ring algorithm for
all-reduce~\cite{thakur2005optimization} is:

\begin{equation}
  T_{\mathit{comm}}(\mathit{batch}) = 2\sum_{i=0}^{L}\left(\alpha \ceil{\log(P)}+ \beta \frac{P-1}{P} \lvert W_i \rvert \right),
  \label{eq:data_comm}
\end{equation}
\noindent where $|W_i|$ is the total number of model parameters in the $i$th
layer. Here, the factor of 2 is merely due to the all-reduce algorithm~\cite{thakur2005optimization}. Note
that for $P \gg 1$ the bandwidth costs are independent of $P$ and unlike the model parallel case does
not depend on the batch size. Most of the current work on distributed training uses batch parallel
to scale training~\cite{goyal2017accurate,you2017imagenet}. The DistBelief paper~\cite{dean2012large} provides easy-to-understand descriptions of model and batch parallelism.

For a convolutional layer, based on Equation~\ref{eq:parameters}, the ratio of communication volume between pure model and batch parallelism becomes
\begin{align}
 \frac{T_{\mathit{comm-volume}}(\mathit{batch})}{T_{\mathit{comm-volume}}(\mathit{model})} &= \frac{ 2 \lvert W_i \rvert }{3 B d_i} 
 					= \frac{(2 k_h k_w X^i_C) Y^i_C}{3 B Y^i_H Y^i_W Y^i_C} \nonumber \\
 											    &=  \frac{2 k_h k_w X^i_C}{3 B Y^i_H Y^i_W } 
\end{align}
												
Consequently, whenever $B > (2 k_h k_w X^i_C / 3 Y^i_H Y^i_W)$, pure batch parallelism is favorable to pure model parallelism. Surprisingly, it is not a foregone conclusion that
batch parallelism is always favorable to model parallelism for convolutional layers. For several convolutional layers
that are used in practice (such as those found in AlexNet with 3x3 filters on 13x13x384 activations), model parallelism has lower communication volume than 
batch parallelism for $B \leq 12$.

If one were to switch from a data parallel distribution shown in Figure~\ref{fig:data} to a model parallel distribution shown in Figure~\ref{fig:model}, the only added
communication cost is the redistribution of $X$ to processes using an all-gather operation, with an associated cost of 

\begin{equation}  T_{\mathit{comm}}(\text{redistribute batch to model}) = \alpha \ceil{\log(P)} + \beta B \frac{P-1}{P} d_i. \end{equation}

It is important to note that this redistribution cost is asymptotically free because the subsequent model parallel step has communication cost that
is three times of the cost of the redistribution.

{\bf Communication costs of pure domain parallelism.} 
A third possibility for parallelization is domain parallel~\cite{jin2018spatially}, where one can decompose the input activation
map as shown in~\fref{fig:domain}. Here each process contains all of the model
parameters (as in the pure batch parallel case), but performs the convolutions
only on a subset of the input image, and writes a subset of the output
activations. For convolutions with filter size larger than one, we have to
perform a halo exchange to communicate the boundary points. This can be
performed as a non-blocking, pair-wise exchange while the convolution is being
applied to the rest of the image. This means that the convolutions that do not
require this boundary data could be computed while the communication is being
performed. The cost of the communication in this case will be:

\begin{equation}
   \begin{split}
     T_{\mathit{comm}}(\mathit{domain}) = &\sum_{i=0}^{L}\left(\alpha + \beta B X^i_WX^i_C\floor{k^i_h/2}\right) \\
                               + &\sum_{i=0}^{L}\left(\alpha + \beta B Y^i_WY^i_C\floor{k^i_w/2}\right) \\
                               +2&\sum_{i=0}^{L}\left(\alpha \ceil{\log(P)}+ \beta \frac{P-1}{P} |W_i|\right),
   \end{split}
  \label{eq:domain_comm}
\end{equation}

\noindent where $X^i_W,\ X^i_H,\ X^i_C, Y^i_W,\ Y^i_H,\ Y^i_C$ are the input/output activation's width, height, and channel size in the $i$th layer, and $k^i_h,\ k^i_w$ is the
corresponding convolution size of that layer. Note that for a $1\times1$ convolution no communication is needed.
For layers with large input activation size and large number of convolution filters, this approach can reduce the computation
time with good strong scaling efficiency. However, it is not effective for small image sizes and not applicable to fully connected layers.

Model parallelism, as published in literature, corresponds to performing a 1D
distribution of the matrix $W_i$, replicating $X_i$ and gathering $Y_i$ multivectors
after multiplication. The $k$th processor can perform its local matrix
multiplication of the form $W_i(k,:) \, X_i$ without any communication, but in
order to fully assemble $Y_i$, each processor needs to gather other
components from other processes. Even if input/output multivectors were also distributed, the
communication bounds stay the same, because while this communication
time would not be necessary for the output $Y_i$, it would be needed for gathering
$X_i$ before the local multiplication.

By contrast, in data parallelism, every process starts with the same parameters, which get updated by the same gradient. 
In fact, the forward pass of batch parallel training needs no communication.
The communication in this case happens during backpropagation, 
where a collective all-reduce operation is needed to compute the total sum of the partial gradients. 
The parallel matrix multiplications in the batch parallel case
are illustrated in Figure~\ref{fig:data}, where the input activations $X_i$ and the output activations $Y_i$ are 
distributed 1D columnwise to processes.

\subsection{Integrated Model and Batch Parallelism}

\begin{figure}[!htbp] 
\begin{subfigure}{1\linewidth} 
  \centering
\includegraphics[width=0.9\textwidth]{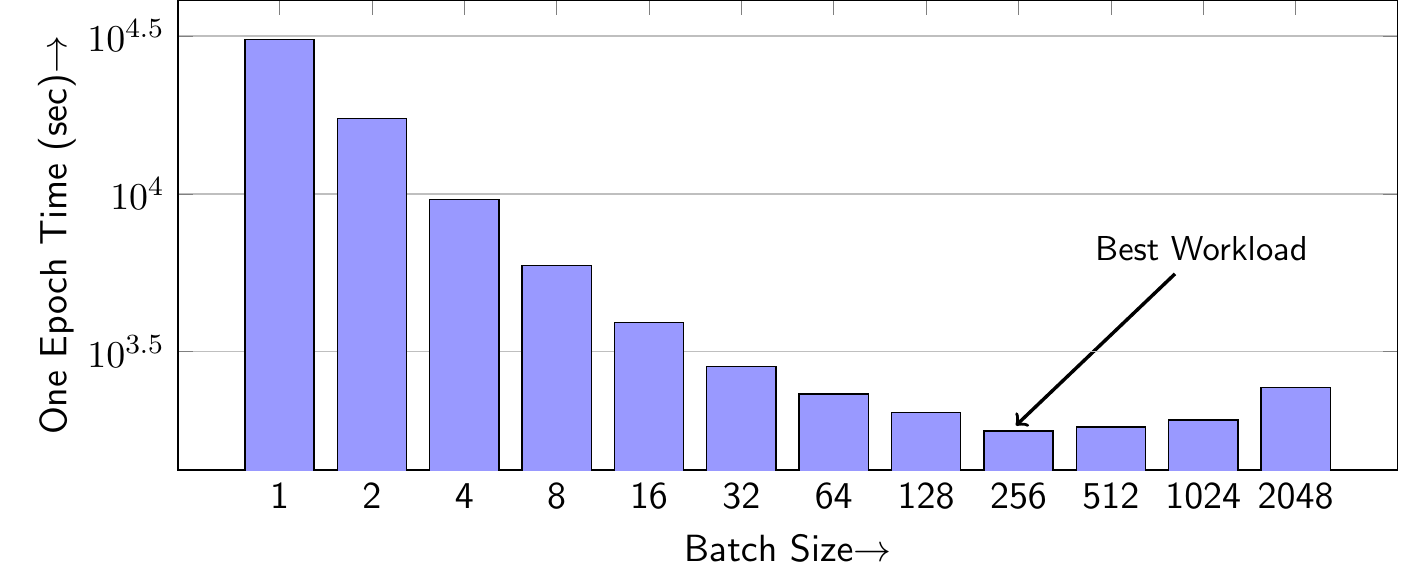}
\end{subfigure}
\caption{One epoch training time of AlexNet computed on a single KNL. Increasing batch size up to 256, reduces the time due to better use of hardware resources
and fewer SGD updates.}
\label{fig:for_bck_time}
\end{figure}

We first discuss the integrated model and batch parallelism and then discuss the full integration
with domain parallelism which extends the scalability limit of the pure batch method. 
Batch parallelism has a favorable communication complexity, but there is an
inherent limit on increasing the batch size. Furthermore, small batch size training
is not efficient in terms of hardware utilization and ultimately training time.
This is due to the fact that small matrix-matrix operations (aka level-3 BLAS operations) cannot
use all the hardware resources, in terms of cores or vectorized units. This is empirically shown in
\fref{fig:for_bck_time}, where we report one epoch training time of AlexNet for different batch sizes measured on a single
Intel Knights Landing (KNL) processor.
The fastest training time is achieved with a batch size of $256$. With the batch parallel
approach one has no choice but to reduce per process batch size for scaling before hitting the limit of 1 batch per process.

\begin{figure}[!htbp]
    \includegraphics[width=0.48  \textwidth]{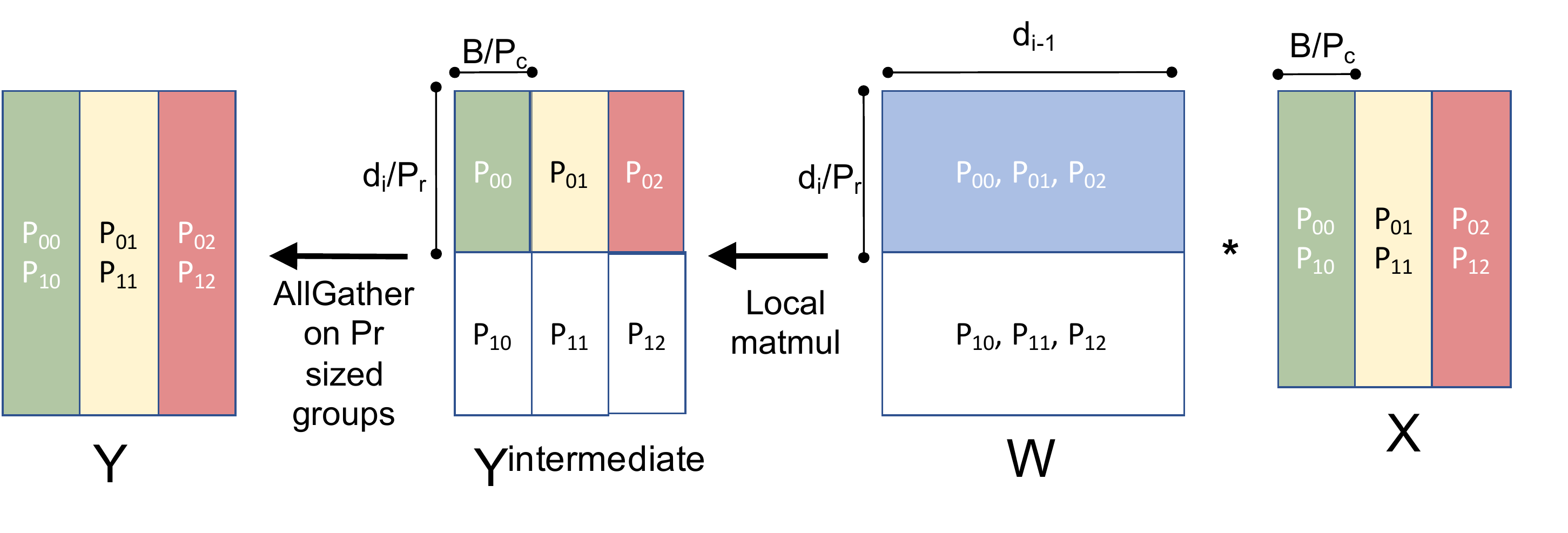}\\\quad
    \includegraphics[width=0.48  \textwidth]{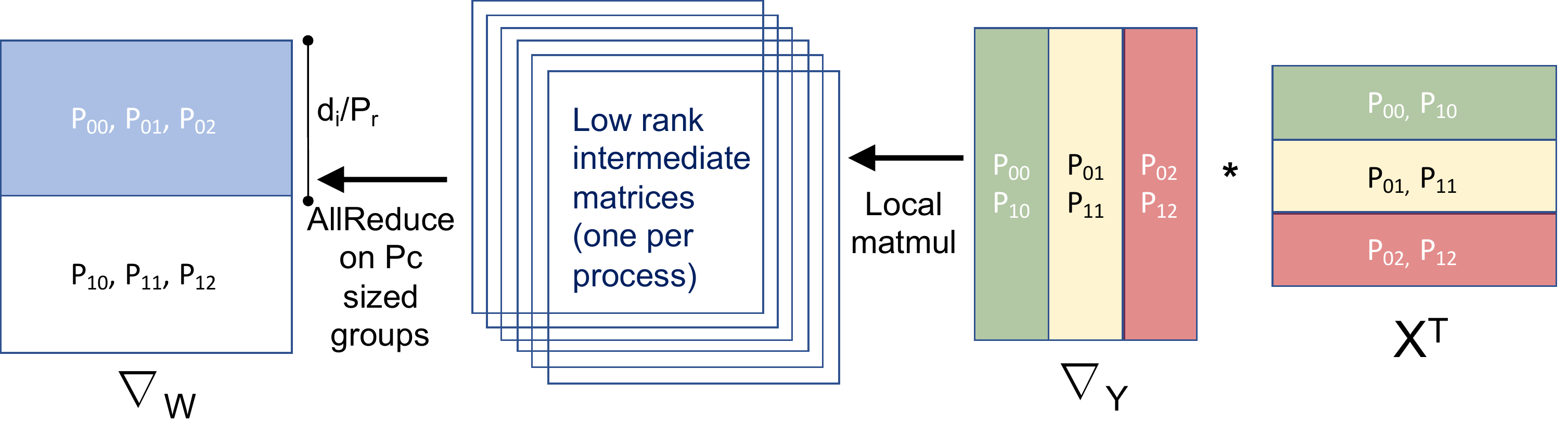}\\\quad
    \includegraphics[width=0.5  \textwidth]{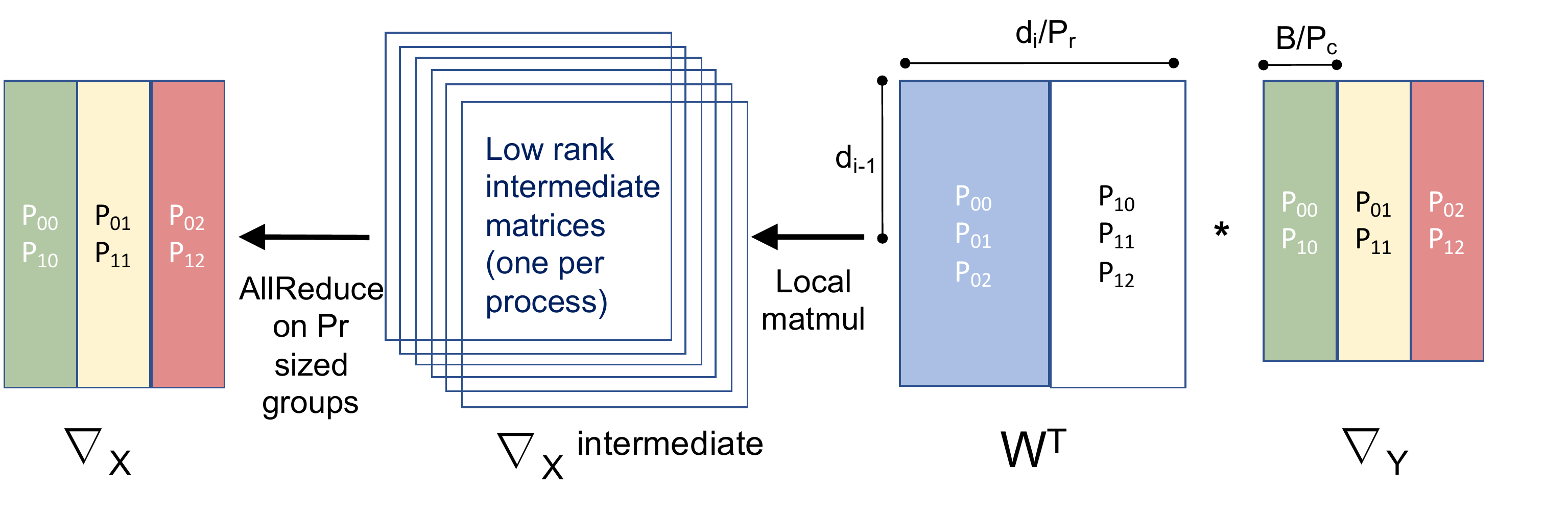}
    \captionof{figure}{1.5D matrix multiply illustration for integrated
    parallel DNN training (top: forward pass, middle/bottom: weight gradient
  computation) using a $2\times3$ process grid indexed as $P_{ij}$.}
    \label{fig:replication} 
\end{figure}

Our {\em integrated batch and model parallel approach} allows us to 
reduce the communication overhead of the pure batch parallel case.
Here, we consider replicating a
subset of $W_i$ as opposed to all of it, a concept that has been explored under
the name of 1.5D algorithms for matrix multiplication~\cite{spdmmm16}. We think
of our process grid logically partitioned as $P = P_r \times P_c$. Each process
holds $(1/P_r)$th piece of $W_i$, effectively replicating $W_i$ matrix $P_c$
times (as opposed to $P$ times in batch parallelism). Conversely, data matrices
are replicated $P_r$ times and each process holds $(1/P_c)$th piece of $X_i$
and $Y_i$.  Communication cost of this 1.5D algorithm, which is illustrated in
Figure~\ref{fig:replication}, is: 

 \begin{equation}
 \begin{split}
   T_{comm} =&  \sum_{i=1}^{L}\left(\alpha \ceil{\log(P_r)} + \beta \frac{B}{P_c} \frac{P_r-1}{P_r} {d_i}\right) \\
   +& 2\sum_{i=2}^{L}\left(\alpha \ceil{\log(P_r)} + \beta \frac{B}{P_c} \frac{P_r-1}{P_r} {d_{i-1}}\right) \\
   +& 2\sum_{i=0}^{L}\left(\alpha \ceil{\log(P_c)} + \beta \frac{P_c-1}{P_c} \frac{|W_i|}{P_r}\right).
   \end{split}
      \label{eq:hybrid}
\end{equation}

Note that unlike in~\eref{eq:data_comm}, the all-reduce communication volume is now reduced by a factor of $P_r$.
This provides a theoretically sound integration of batch and model parallelism.
It can be especially valuable 
for networks with many fully connected layers.
Furthermore, this algorithm
automatically selects the best configuration to distribute the model and batch parallel work
given a fixed batch size on $P$ processes. The closest approach to ours is the hybrid model/batch
parallel approach described by Das et al.~\cite{das2016distributed}, but that paper does not describe the details 
of the partitioning of the data and the model to the processes. In addition, the authors claim that
using any other dimension to extract parallelism would always be sub-optimal, which we show not be true in general 
by using domain parallelism. 

Similar to the analysis of pure model and pure batch cases, the cost of redistribution is asymptotically amortized in this integrated batch and model parallel case as well. 
In particular, if one were to switch process grids in between layers, say from a pure batch case ($1\times P$ grid) to a balanced case ($\sqrt{p} \times \sqrt{p}$) grid, 
the communication costs would asymptotically stay constant.

For the curious reader familiar with the theory of parallel matrix
multiplication, we would like to clarify why we consider our approach a 1.5D
algorithm, as opposed to a 2D algorithm such as Cannon's
algorithm~\cite{cannon} or SUMMA~\cite{van1997summa}.  2D matrix multiplication
algorithms are optimal in terms of their memory usage; that is, each processor
only holds $(1/p)$th of the total memory needed to store all three matrices (2
inputs and 1 output). In other words, there is no replication. The class of .5D
algorithms (of which 1.5D algorithm is a member), by contrast, are not optimal
in terms of memory consumption. At least one matrix is replicated multiple
times, which often results in an asymptotic reduction in communication
costs~\cite{ballard2011minimizing}. This is indeed the case for the algorithm
described in Figure~\ref{fig:replication}.

\begin{figure*}[!htbp]
\begin{subfigure}{0.42\linewidth}
  \centering
\includegraphics[width=.84\textwidth]{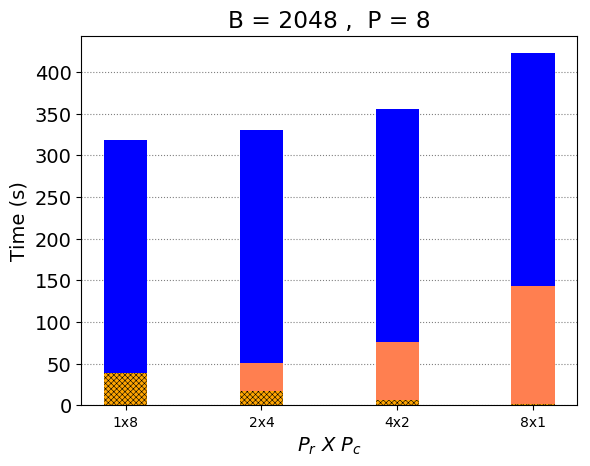}
\end{subfigure}
\begin{subfigure}{0.42\linewidth}
  \centering
\includegraphics[width=.84\textwidth]{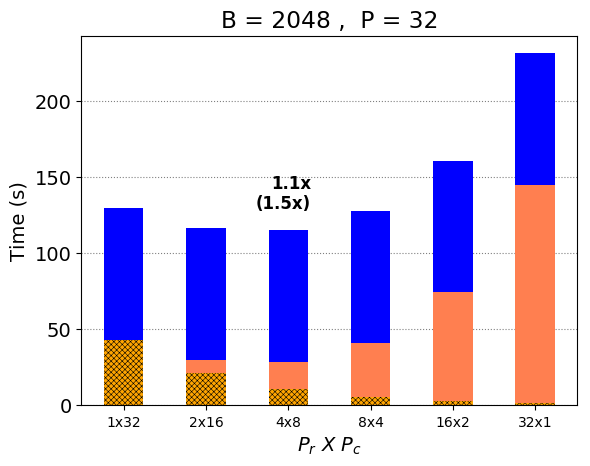}
\end{subfigure}
\begin{subfigure}{0.42\linewidth}
  \centering
\includegraphics[width=.84\textwidth]{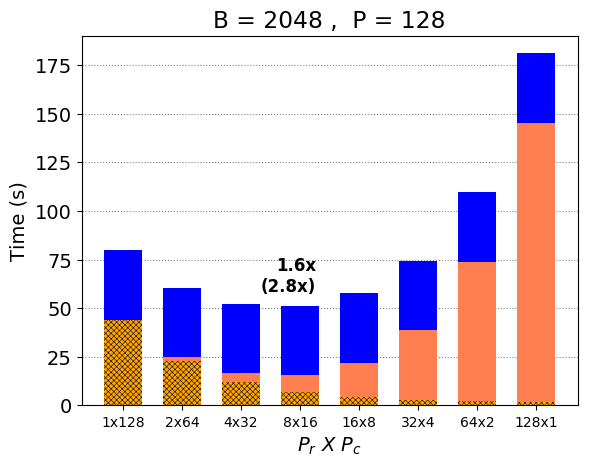}
\end{subfigure}
\begin{subfigure}{0.42\linewidth}
  \centering
\includegraphics[width=.84\textwidth]{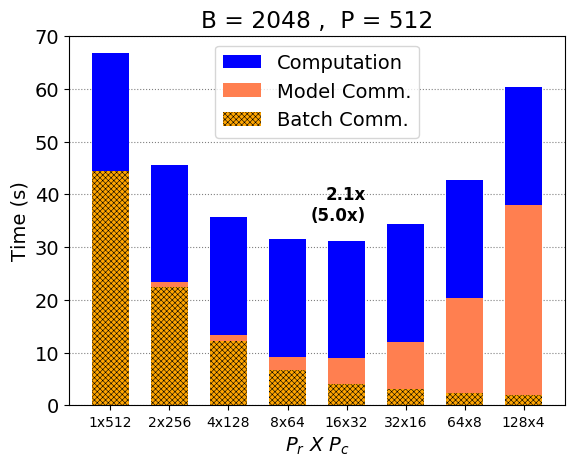}
\end{subfigure}
\caption{
  Strong scaling analysis of integrated model and batch parallel approach using the simulated results.
  The orange bar shows the total communication time, with the cross hatched portion representing
  the time spent in batch parallel communication (i.e. the ring all-reduce during backprop). 
  Here we use the same process grid for all layers, which means some amount of model parallelism is used for both convolutional and FC layers when $P_r>1$.
  The speedup for the total time compared to pure batch parallel is shown in bold text on top of the best bar chart.
  We also report the corresponding speedup for communication time in parenthesis. 
  In strong scaling, we keep the global batch size fixed, and increase the number of processes to reduce the training time. 
 }
\label{fig:strong_batch_model_all}
\end{figure*}

\subsection{Integrated Model, Batch and Domain Parallelism}
The pure batch parallel method has a theoretical strong scaling limit of $B$. In the limit each
process gets a batch size of one (i.e.\ it reads a single data). It is possible to extend this limit with the
integrated model and batch parallel approach discussed above. But this approach is sub-optimal for early layers
of the network, as the all-gather communication volume is very high there (\eref{eq:hybrid}). This is due to the fact that this communication
volume depends on the size of the activation map (i.e. $Y_i$) which is prohibitively large in the beginning layers.


\begin{figure*}[!htbp]
\begin{subfigure}{0.42\linewidth}
  \centering
  \includegraphics[width=.84\textwidth]{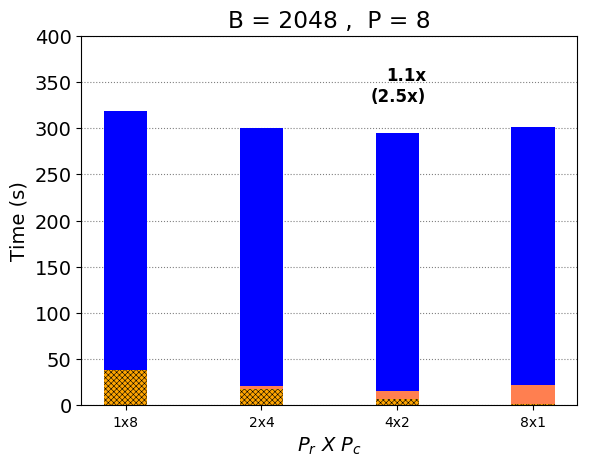}
\end{subfigure}
\begin{subfigure}{0.42\linewidth}
  \centering
  \includegraphics[width=.84\textwidth]{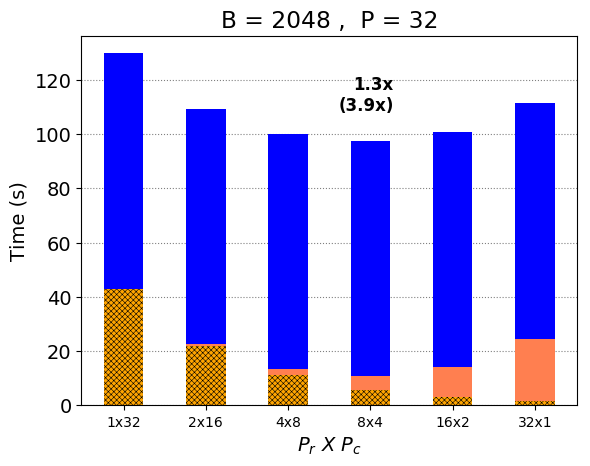}
\end{subfigure}
\begin{subfigure}{0.42\linewidth}
  \centering
  \includegraphics[width=.84\textwidth]{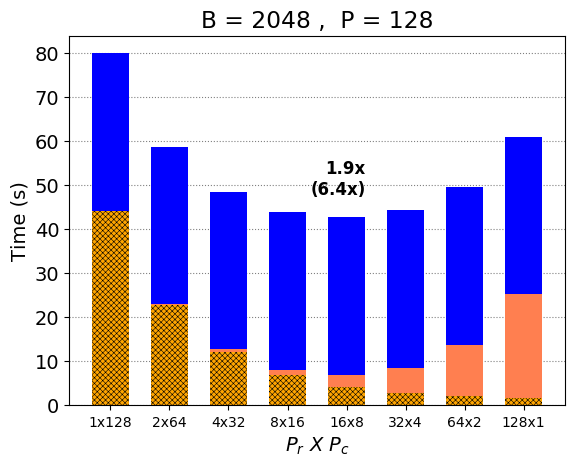}
\end{subfigure}
\begin{subfigure}{0.42\linewidth}
  \centering
  \includegraphics[width=.84\textwidth]{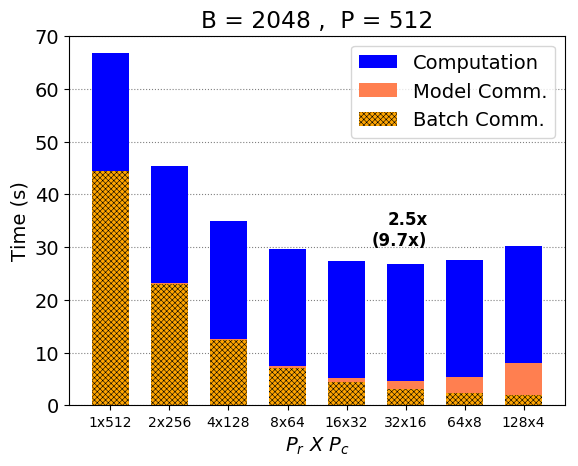}
\end{subfigure}
\caption{
   Strong scaling analysis of integrated model and batch parallel approach using the simulated results.
   Model parallelism is used in FC layers only.
   Notice the significant improvement in best time compared to~\fref{fig:strong_batch_model_all} which uses model parallelism in both convolutional and FC layers.
}
\label{fig:strong_batch_model_FC}
\end{figure*}

However, as we show below the domain parallel approach has a favorable communication complexity for early
layers of a neural network where the input activation size is large. For these
layers it is favorable to use domain parallelism instead of model parallelism, as
it leads to a smaller communication volume that can actually be overlapped with part of the computation in both forward and backward pass.
Note that in model parallel
one has to perform a blocking all-gather operation which is detrimental for
performance.
Moreover, the domain parallel approach does not require any communication for $1\times 1$ convolutions which are actually becoming
a dominant portion of the network in recent architectures~\cite{he2016deep}.
However, for fully connected layers the halo exchange region will consist of all of the input activations. To avoid that large communication
cost we can actually integrate all the three parallelism methods.
The communication
complexity for integrating all the three methods would then become:

 \begin{equation}
   \begin{split}
         T_{comm} = &\sum_{i\in L_M}\left(\alpha \ceil{\log(P_r)} + \beta \frac{B}{P_c} \frac{P_r-1}{P_r} {d_i}\right) +\\
                    2&\sum_{i\in L_M}\left(\alpha \ceil{\log(P_r)} + \beta \frac{B}{P_c} \frac{P_r-1}{P_r} {d_{i-1}}\right) +\\
                    2&\sum_{i\in L_M}\left(\alpha \ceil{\log(P_c)} + \beta \frac{P_c-1}{P_c} \frac{|W_i|}{P_r}\right)+\\
                     &\sum_{i\in L_D}\left(\alpha + \beta \frac{B}{P_c}X^i_WX^i_C\floor{k^i_h/2}\right) + \\
                     &\sum_{i\in L_D}\left(\alpha + \beta \frac{B}{P_c}X^{i+1}_WX^{i+1}_C\floor{k^i_w/2}\right) + \\
                    2&\sum_{i\in L_D}\left(\alpha \ceil{\log(P)} + \beta \frac{P-1}{P} {|W_i|}\right),
   \end{split}
   \label{eq:hybrid_domain}
\end{equation}

\noindent where $L_M$ and $L_D$ refer to the list of layers where the $P_r$
groups are used to partition either the model or the domain.  Note that for
$L_M=L$, $L_D=0$, we get the integrated model and batch parallel complexity as
expected.

The choice of whether to partition the model or the domain can be made by
computing the communication complexity.  Generally, it is better to use domain
parallelism for the initial layers of the network, since the activation size
is large.  However, the domain parallel approach loses its communication advantage for fully connected
layers (for which $k_h=X_H,\ k_w=X_W$).

\section{Simulated Performance in Training AlexNet}
{\bf Simulation setup.} We analytically explore the spectrum of both the integrated
batch and model parallel approach, as well as the full integration with domain parallelism by simulating Eq.~\ref{eq:hybrid} and Eq.~\ref{eq:hybrid_domain}.
To limit the number of variables, we fix a network (AlexNet), a training set of images
(ImageNet LSVRC-2012 contest), and a computing platform (NERSC's Cori
supercomputer). These fixed options, described in
Table~\ref{tab:simulation_fixed}, are chosen just to develop a proof-of-concept
of our integrated batch and model parallel approach. 

We considered two scenarios: (a) $B\geq P$: here the relevant integration is between model
and batch parallel approaches and domain parallelism is not used as its communication overhead
is higher than batch parallel (Eq.~\ref{eq:domain_comm})
(b) $B<P$:  This is the case where we reach the maximum scaling limit of the batch parallel method, and use domain
parallelism to scale beyond this
(Eq.~\ref{eq:hybrid_domain}).  
For the first scenario, we considered two cases.
At first, the same process grid is used for all layers of the network, which means that if $P_r> 1$ then some amount of model parallelism will be used even in convolutional layers.
Then we considered the improved case where we force $P_r=1,P_c=P$ for the convolutional layers and use varying $P_r\times P_c$ grids for the fully connected layers.

We compute the communication time for a single
iteration with various choices of the mini-batch size $B$, the number of
processes, and the configuration of process grid $P_r\times P_c$. Using this data, we then
compute the communication time for a complete epoch by multiplying the
communication time form Eq.~\ref{eq:data_comm} by $N/B$. A typical simulation of the Neural Network would require
many epochs of training (100 epochs in the case of AlexNet~\cite{krizhevsky2012imagenet}).

Furthermore, we also consider the computational time by empirically measuring the time needed for an
SGD iteration for AlexNet on a single KNL using Intel Caffe as shown in~\fref{fig:for_bck_time}.
We use this data for cases with the same computational workload to compute the total run time.


 \begin{table}[!t]
    \centering
    \begin{tabular}{@{} lll @{}} 
       \toprule       
          &  Fixed options & Relevant parameters \\
       \midrule
        Network     & AlexNet~\cite{krizhevsky2012imagenet} & 5 convolutional and  \\
         architecture     & parameters: 61M &  3 fully connected layers \\
                  \midrule
       Training       & ImageNet   & training images:  1.2M\\
         images &  LSVRC-2012 contest & Number of categories: 1000 \\
        \midrule
       Computing 	       &  \multirow{ 3}{*}{NERSC's Cori2}   &  Processor: Intel
KNL
\\
platform &   &  latency: $\alpha = 2\mu$ s  \\
 &   &  inverse bw:  $1/\beta = 6$GB/s  \\
       \bottomrule
    \end{tabular}
    \caption{
    Fixed parameters used to simulate the cost of training neural networks
  using integrated batch and model parallel approach. We only change the
mini-batch size and the number and configurations of processes in the presented
results.}

    \label{tab:simulation_fixed}
 \end{table}

{\bf Strong scaling with a fixed mini-batch size.}
At first, we present the strong scaling results for integrated model and batch.
We initially apply the integrated method in a way that the same process grid is used for all layers of the network, which means that if $P_r>1$ then some amount of model parallelism will be used even in convolutional layers.
The results are shown~\fref{fig:strong_batch_model_all} where the training was performed using
$P=8$ to $P=512$ processes with a fixed mini-batch size of $B=2,048$.  In each
subfigure in~\fref{fig:strong_batch_model_all}, only the configurations of the process
grid vary.
We can see that even in the naive format, better performance can be attained
with an integrated batch and model parallelism, especially for larger values of $P$. 
For example, on $P=512$ processes, the best performance is observed
with $16\times32$ process grid which results in $2.1\times$ speed up in the overall runtime and
$5.0\times$ speedup in communication (
Figure~\ref{fig:strong_batch_model_all}-d).  The improved performance is
primarily driven by reduced communication needed by the integrated model and
batch parallel approach (notice the reduction of the communication volume of the parameters by $P_r$ factor in Eq.~\ref{eq:hybrid}).
However, the benefit of the
integrated approach is not realized on a relatively small number processors, such as
with 8 processes in Figure~\ref{fig:strong_batch_model_all}(a).  The first reason
is that here the main bottleneck is computation. Moreover, the communication time for
model parallel does not scale down 
since per process batch size is very large (note the $B/P_c$ term in~Eq.\ref{eq:hybrid}).



\begin{figure}[!htbp]
   \centering
   \includegraphics[width=.84\linewidth]{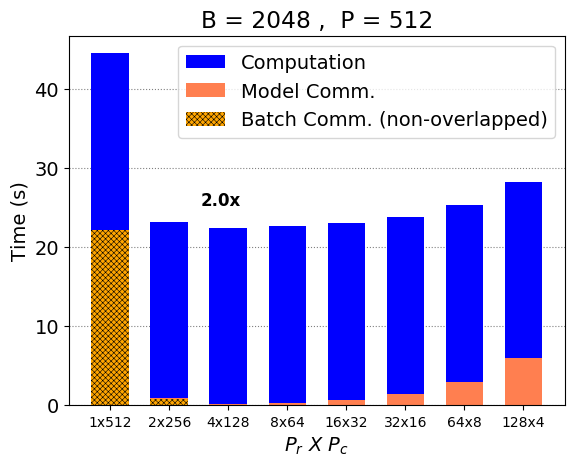} 
   \caption{Here we show results for perfect overlapping of communication with backpropagation part of the computations.}
   \label{fig:overlap_comm}
\end{figure}

Next, we considered the improved case where we force $P_r=1,P_c=P$ for the convolutional layers and use varying $P_r\times P_c$ grids for the fully connected layers.
For the configurations considered, this results in using pure batch parallelism in convolutional
layers and both model and batch parallelism in FC layers as shown in
Figure~\ref{fig:strong_batch_model_FC}.  Making the convolutional layer pure
batch parallel can reduce the communication significantly, as evident by
comparing~\fref{fig:strong_batch_model_FC}
and~\fref{fig:strong_batch_model_all}. 
For instance, the case with $B=2048,\ P=512$ results in $2.5\times$ speedup in 
overall runtime and $9.7\times$ speedup in communication time (Figure~\ref{fig:strong_batch_model_FC}-d).
We also show how the results would change if we consider a perfect overlap between communication and computation as shown in~\fref{fig:overlap_comm}.
This overlapping can only be performed with the backpropagation phase, where the all-reduce communication can happen
while the transpose convolution of next layers are being performed (which accounts for two-thirds of the communication).
Even in this setting there is $2.0\times$ speedup. 
We believe that this speed up is actually going to increase, 
given the new domain specific architectures optimized for accelerating the computation
part of neural network training/inference.

\begin{figure*}[!htbp] 
\begin{subfigure}{0.42\linewidth}
  \centering
\includegraphics[width=.84\textwidth]{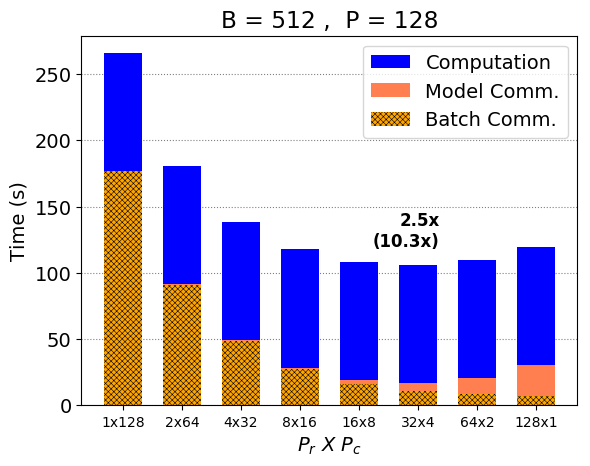}
\end{subfigure}
\begin{subfigure}{0.42\linewidth}
  \centering
\includegraphics[width=.84\textwidth]{new_python_code/Comm_Comp_B=2048,P=512.png}
\end{subfigure}
\begin{subfigure}{0.42\linewidth}
  \centering
\includegraphics[width=.84\textwidth]{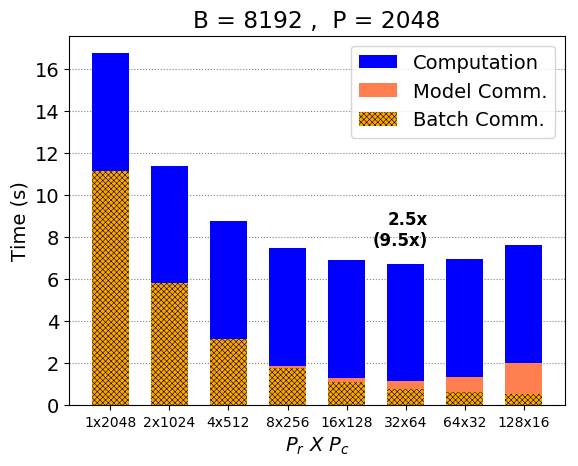}
\end{subfigure}
\begin{subfigure}{0.42\linewidth}
  \centering
\includegraphics[width=.84\textwidth]{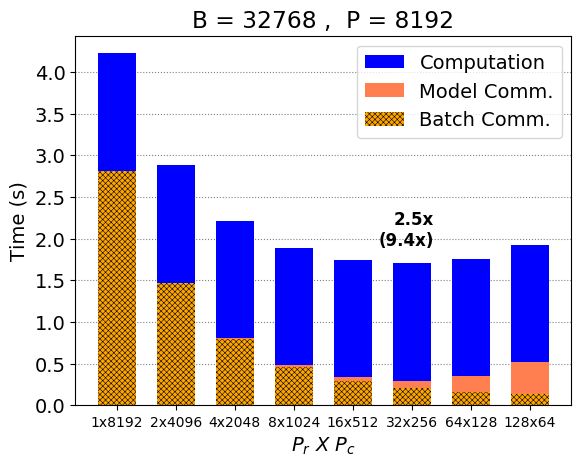}
\end{subfigure}
\caption{
  Simulated cost of integrated model and batch parallel method. We present weak scaling results for the communication and computation complexity
  when training AlexNet. 
  The speedup for the total time compared to pure batch parallel is shown in bold text on top of the best bar chart.
  We also report the corresponding speedup for communication time in parenthesis. 
	Here we use the same process grid for all layers, which results in using some amount of model parallelism (when $P_r>1$) for convolutional layers as well, which is sub-optimal. A better
  approach is to use pure batch parallelism for convolutional layers.
}
    \label{fig:weak_batch_model_FC}  
\end{figure*}

{\bf Scaling with a variable mini-batch size.}
We now consider weak scaling by varying the mini-batch size and the process grid simultaneously,
as shown in~\fref{fig:weak_batch_model_FC}. Here we use choose model/batch parallel based on the complexity analysis
of Eq.~\ref{eq:hybrid} (similar to the strong scaling shown in~\fref{fig:strong_batch_model_FC}).
In each subfigure, only the configurations of the process grid vary for a fixed $P$ and
$B$.  Similar to the strong scaling results, we observe that the integrated
approach can reduce the communication significantly as we change the mini-batch
size.



\begin{figure*}[!htbp] 
\begin{subfigure}{0.42\linewidth}
  \centering
\includegraphics[width=.84\textwidth]{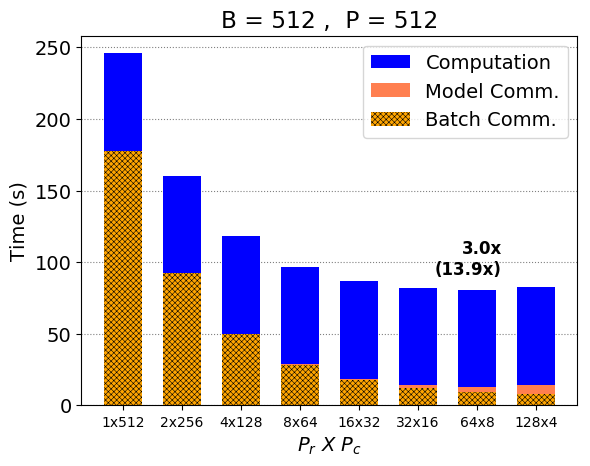}
\end{subfigure}
\begin{subfigure}{0.42\linewidth}
  \centering
\includegraphics[width=.84\textwidth]{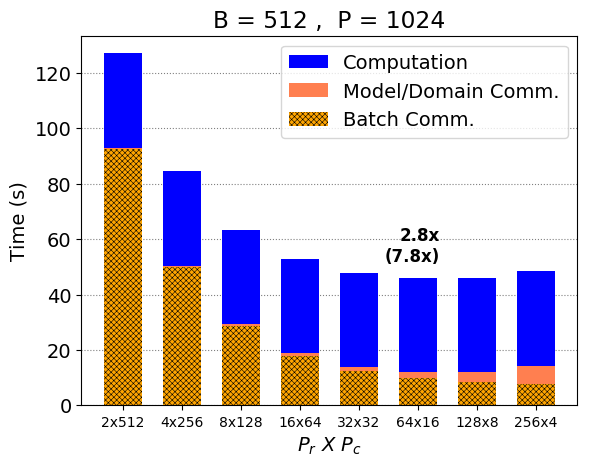}
\end{subfigure}
\begin{subfigure}{0.42\linewidth}
  \centering
\includegraphics[width=.84\textwidth]{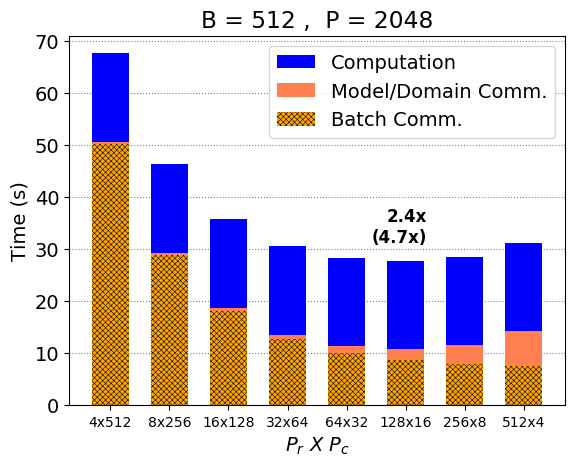}
\end{subfigure}
\begin{subfigure}{0.42\linewidth}
  \centering
\includegraphics[width=.84\textwidth]{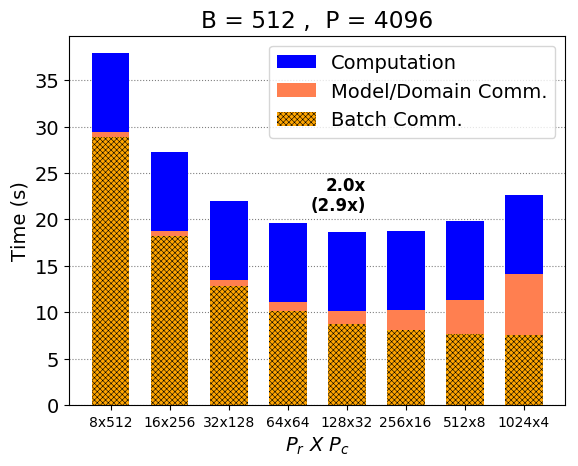}
\end{subfigure}
\caption{
 Illustration of how domain parallel can extend strong scaling limit of pure batch parallelism.
}
\label{fig:strong_domain_parallel}
\end{figure*}


{\bf Scaling beyond batch size.}
The pure batch parallel method has a scaling limit to the maximum batch size that one can use.
However, one cannot increase batch size indefinitely as it is known to be detrimental to the performance
of the Neural Network~\cite{keskar2016large}. Recent works have tried to increase this limit by changing
the hyper-parameters of SGD~\cite{goyal2017accurate,you2017scaling}, but these methods also hit a limit and have been
only shown to work for certain applications in ImageNet classification. So a natural question is how do we scale beyond 
this theoretical limit with pure batch parallelism? One could use the integrated approach and scale the model part for all layers, but
as shown above this results in sub-optimal communication time. A better approach is to use an integrated batch, domain, and model
parallel where for the initial layers we use domain parallel instead of the model. Note that the domain parallel approach requires
a much smaller communication as compared to model parallel, and actually requires no communication for $1\times 1$ convolutions (Eq.~\ref{eq:domain_comm}).
To illustrate this, we show the scaling results for $B=512$ up to $P=4096$ in~\fref{fig:strong_domain_parallel}.
In~\fref{fig:strong_domain_parallel}(a), convolutional layers use pure batch parallelism with per-process batch size set to one.
By contrast, in~\fref{fig:strong_domain_parallel}(c-d), each image is partitioned into 2,4, and 8 parts where each process works with one part of the image.
Using this integrated batch, domain, and model parallel approach, we can continue scaling beyond the theoretical limit with pure batch parallelism (beyond 512 processes in~\fref{fig:strong_domain_parallel}).

\section{Discussion}

One disadvantage of batch parallelism over model and domain parallelism is that it tends to change the convergence characteristics of 
DNN training algorithms as larger minibatches beyond a certain point can hurt accuracy. Our integrated framework also provides guidance
on how to choose the right parallelization parameters if the user decides to limit the maximum allowable batch parallelism in light of accuracy concerns related to large batch sizes.

Due to DNN training being computational intensive, memory considerations have been secondary to performance. Solutions that exploit pure data parallelism often replicate the whole model in each node. By contrast, the 1.5D matrix-multiplication algorithms used by our integrated parallel approach cut down the model replication cost by a factor of $p_r$, at the cost of an increase in data replication by a factor of $p_c$. Like our communication costs, our memory costs are simply a linear combination of the memory costs of these two extremes of pure data and 
pure model parallelism. 

We also considered the alternative of using 2D matrix multiplication algorithms instead of the 1.5D algorithm. The popular stationary-C variant of the 2D SUMMA algorithm~\cite{van1997summa} is symmetrical in nature; in the sense that it communicates equal proportions of both input matrices for an operation $C=A B$. When matrices $A$ and $B$ are of comparable sizes, this is a good fit. Often in deep learning, one of the matrices is bigger than the other. For such situations, there are other less-common variants of SUMMA that keep another matrix stationary~\cite{schatz2016parallel}. These algorithms are more complicated that our 1.5D algorithm, and communicate more data either asymptotically or by higher constants. 

Consider stationary-A SUMMA, which is the best fit for the forward propagation $Y = W X$ among all 2D algorithm variants. This algorithm has $4$ communication steps compared to a single step in our algorithm. For simplicity assume that $d_i = d_{i-1}$. Also assume that $p_r$ and $p_c$ are large enough such that $(p_r-1)/p_r \approx (p_c-1)/p_c  \approx 1$. When $\lvert W_i \rvert > B  d_i$, it communicates $2 B \, d_i / p_r + B \, d_i /p_c$ words, compared to our 1.5D algorithm's $B \, d_i /p_c$ words. In that sense, its communication costs approach 1.5D when $p_r \gg p_c$ but never surpass it. When $\lvert W_i \rvert < B  d_i$, all possible 2D algorithms become asymptotically slower because they have to communicate two matrices and no matter which two they choose, the costs become higher than solely communicating the single smaller matrix. By contrast, our 1.5D algorithm communicates only that single matrix. Hence, there is no regime where 2D becomes strictly favorable in terms of communication volume. The main advantage of 2D algorithms over 1.5D algorithm is that their memory consumption is optimal in the sense that they do not perform any asymptotic data replication. Memory consumption optimality might be a legitimate concern depending on the platform and the DNN model size.

\section{Conclusion}
We presented an integrated parallel algorithm that exploits model, batch, and
domain parallelism in training deep neural networks (DNNs). 
We discussed the associated communication complexity by
analyzing both forward and backwards pass, and showed that theoretically the
integrated parallel approach can achieve better run time. Furthermore,
the integrated parallel approach increases the scalability limit of the
pure batch parallel method that is commonly used, by decomposing both along the weight matrix
as well as the domain. This approach allows optimal selection of per
process batch size and model size which results in better throughput as
compared to pure batch/model parallel algorithms. 

Our analysis toolset is 
primarily comprised of parallel matrix algorithms. In particular, the analysis of our integrated model and batch parallel approach relies on a communication-avoiding 
1.5D matrix multiplication algorithm. This explicit connection between parallel matrix algorithms and 
DNN training has the potential to enable
the discovery of new classes of parallel algorithms and lower bounds for training DNNs.

\section{Acknowlegments}

This manuscript has been authored by an author at Lawrence Berkeley National Laboratory under Contract No. DE-AC02-05CH11231 with the U.S. Department of Energy. The U.S. Government retains, and the publisher, by accepting the article for publication, acknowledges, that the U.S. Government retains a non-exclusive, paid-up, irrevocable, world-wide license to publish or reproduce the published form of this manuscript, or allow others to do so, for U.S. Government purposes.

This work was supported by the Laboratory Directed Research and Development Program of Lawrence Berkeley National Laboratory under U.S. Department of Energy Contract No. DE-AC02-05CH11231.

The authors would also like to acknowledge generous support from Intel's VLAB team for providing access to KNLs.

\bibliographystyle{plain}
\bibliography{ref}

\begin{thebibliography}{10}

\bibitem{ballard2011minimizing}
Grey Ballard, James Demmel, Olga Holtz, and Oded Schwartz.
\newblock Minimizing communication in numerical linear algebra.
\newblock {\em SIAM Journal on Matrix Analysis and Applications},
  32(3):866--901, 2011.

\bibitem{cannon}
Lynn~Elliot Cannon.
\newblock {\em A cellular computer to implement the {Kalman} filter algorithm}.
\newblock PhD thesis, Montana State University, 1969.

\bibitem{chan2007collective}
Ernie Chan, Marcel Heimlich, Avi Purkayastha, and Robert Van De~Geijn.
\newblock Collective communication: theory, practice, and experience.
\newblock {\em Concurrency and Computation: Practice and Experience},
  19(13):1749--1783, 2007.

\bibitem{chilimbi2014project}
Trishul~M Chilimbi, Yutaka Suzue, Johnson Apacible, and Karthik Kalyanaraman.
\newblock Project adam: Building an efficient and scalable deep learning
  training system.
\newblock In {\em OSDI}, volume~14, pages 571--582, 2014.

\bibitem{coates2013deep}
Adam Coates, Brody Huval, Tao Wang, David Wu, Bryan Catanzaro, and Ng~Andrew.
\newblock Deep learning with {COTS} {HPC} systems.
\newblock In {\em International Conference on Machine Learning}, pages
  1337--1345, 2013.

\bibitem{das2016distributed}
Dipankar Das, Sasikanth Avancha, Dheevatsa Mudigere, Karthikeyan Vaidynathan,
  Srinivas Sridharan, Dhiraj Kalamkar, Bharat Kaul, and Pradeep Dubey.
\newblock Distributed deep learning using synchronous stochastic gradient
  descent.
\newblock {\em arXiv preprint arXiv:1602.06709}, 2016.

\bibitem{dean2012large}
Jeffrey Dean, Greg Corrado, Rajat Monga, Kai Chen, Matthieu Devin, Mark Mao,
  Andrew Senior, Paul Tucker, Ke~Yang, Quoc~V Le, et~al.
\newblock Large scale distributed deep networks.
\newblock In {\em Advances in neural information processing systems}, pages
  1223--1231, 2012.

\bibitem{goyal2017accurate}
Priya Goyal, Piotr Doll{\'a}r, Ross Girshick, Pieter Noordhuis, Lukasz
  Wesolowski, Aapo Kyrola, Andrew Tulloch, Yangqing Jia, and Kaiming He.
\newblock Accurate, large minibatch {SGD}: Training {ImageNet} in 1 hour.
\newblock {\em arXiv preprint arXiv:1706.02677}, 2017.

\bibitem{havaei2017brain}
Mohammad Havaei, Axel Davy, David Warde-Farley, Antoine Biard, Aaron Courville,
  Yoshua Bengio, Chris Pal, Pierre-Marc Jodoin, and Hugo Larochelle.
\newblock Brain tumor segmentation with deep neural networks.
\newblock {\em Medical image analysis}, 35:18--31, 2017.

\bibitem{he2016deep}
Kaiming He, Xiangyu Zhang, Shaoqing Ren, and Jian Sun.
\newblock Deep residual learning for image recognition.
\newblock In {\em Proceedings of the IEEE conference on computer vision and
  pattern recognition}, pages 770--778, 2016.

\bibitem{jin2018spatially}
Peter Jin, Boris Ginsburg, and Kurt Keutzer.
\newblock Spatially parallel convolutions.
\newblock {\em ICLR 2018 Workshop}, 2018.

\bibitem{jin2016scale}
Peter~H Jin, Qiaochu Yuan, Forrest Iandola, and Kurt Keutzer.
\newblock How to scale distributed deep learning?
\newblock {\em arXiv preprint arXiv:1611.04581}, 2016.

\bibitem{keskar2016large}
Nitish~Shirish Keskar, Dheevatsa Mudigere, Jorge Nocedal, Mikhail Smelyanskiy,
  and Ping Tak~Peter Tang.
\newblock On large-batch training for deep learning: Generalization gap and
  sharp minima.
\newblock {\em arXiv preprint arXiv:1609.04836}, 2016.

\bibitem{kim2016accurate}
Jiwon Kim, Jung Kwon~Lee, and Kyoung Mu~Lee.
\newblock Accurate image super-resolution using very deep convolutional
  networks.
\newblock In {\em Proceedings of the IEEE Conference on Computer Vision and
  Pattern Recognition}, pages 1646--1654, 2016.

\bibitem{spdmmm16}
Penporn Koanantakool, Ariful Azad, Ayd{\i}n Bulu\c{c}, Dmitriy Morozov,
  Sang-Yun Oh, Leonid Oliker, and Katherine Yelick.
\newblock Communication-avoiding parallel sparse-dense matrix-matrix
  multiplication.
\newblock In {\em Proceedings of the IPDPS}, 2016.

\bibitem{krizhevsky2012imagenet}
Alex Krizhevsky, Ilya Sutskever, and Geoffrey~E Hinton.
\newblock Imagenet classification with deep convolutional neural networks.
\newblock In {\em Advances in neural information processing systems}, pages
  1097--1105, 2012.

\bibitem{long2015fully}
Jonathan Long, Evan Shelhamer, and Trevor Darrell.
\newblock Fully convolutional networks for semantic segmentation.
\newblock In {\em Proceedings of the IEEE Conference on Computer Vision and
  Pattern Recognition}, pages 3431--3440, 2015.

\bibitem{brats}
A.~Mang, S.~Tharakan \textmd{A. Gholami}, N.~Himthani, S.~Subramanian,
  J.~Levitt, M.~Azmat, K.~Scheufele, M.~Mehl, C.~Davatzikos, B.~Barth, and
  G.~Biros.
\newblock {SIBIA-GlS}: Scalable biophysics-based image analysis for glioma
  segmentation.
\newblock {\em The multimodal brain tumor image segmentation benchmark (BRATS),
  MICCAI}, 2017{.}

\bibitem{recht2011hogwild}
Benjamin Recht, Christopher Re, Stephen Wright, and Feng Niu.
\newblock Hogwild: A lock-free approach to parallelizing stochastic gradient
  descent.
\newblock In {\em Advances in neural information processing systems}, pages
  693--701, 2011.

\bibitem{ren2015faster}
Shaoqing Ren, Kaiming He, Ross Girshick, and Jian Sun.
\newblock Faster {R-CNN}: Towards real-time object detection with region
  proposal networks.
\newblock In {\em Advances in neural information processing systems}, pages
  91--99, 2015.

\bibitem{rogers1998using}
RO~Rogers and David~B Skillicorn.
\newblock Using the {BSP} cost model to optimise parallel neural network
  training.
\newblock {\em Future Generation Computer Systems}, 14(5):409--424, 1998.

\bibitem{schatz2016parallel}
Martin~D Schatz, Robert~A Van~de Geijn, and Jack Poulson.
\newblock Parallel matrix multiplication: A systematic journey.
\newblock {\em SIAM Journal on Scientific Computing}, 38(6):C748--C781, 2016.

\bibitem{szegedy2015going}
Christian Szegedy, Wei Liu, Yangqing Jia, Pierre Sermanet, Scott Reed, Dragomir
  Anguelov, Dumitru Erhan, Vincent Vanhoucke, and Andrew Rabinovich.
\newblock Going deeper with convolutions.
\newblock In {\em Proceedings of the IEEE conference on computer vision and
  pattern recognition}, pages 1--9, 2015.

\bibitem{thakur2005optimization}
Rajeev Thakur, Rolf Rabenseifner, and William Gropp.
\newblock Optimization of collective communication operations in {MPICH}.
\newblock {\em The International Journal of High Performance Computing
  Applications}, 19(1):49--66, 2005.

\bibitem{van1997summa}
Robert~A Van De~Geijn and Jerrell Watts.
\newblock {SUMMA}: Scalable universal matrix multiplication algorithm.
\newblock {\em Concurrency-Practice and Experience}, 9(4):255--274, 1997.

\bibitem{wu2016squeezedet}
Bichen Wu, Forrest Iandola, Peter~H Jin, and Kurt Keutzer.
\newblock Squeezedet: Unified, small, low power fully convolutional neural
  networks for real-time object detection for autonomous driving.
\newblock {\em arXiv preprint arXiv:1612.01051}, 2016.

\bibitem{squeezeseg}
Bichen Wu, Alvin Wan, Xiangyu Yue, and Kurt Keutzer.
\newblock Squeezeseg: Convolutional neural nets with recurrent crf for
  real-time road-object segmentation from 3d lidar point cloud.
\newblock In {\em In Review}, 2017.

\bibitem{you2017scaling}
Yang You, Igor Gitman, and Boris Ginsburg.
\newblock Scaling {SGD} batch size to 32k for {ImageNet} training.
\newblock {\em arXiv preprint arXiv:1708.03888}, 2017.

\bibitem{you2017imagenet}
Yang You, Zhao Zhang, C~Hsieh, James Demmel, and Kurt Keutzer.
\newblock {ImageNet} training in minutes.
\newblock {\em CoRR, abs/1709.05011}, 2017.

\bibitem{zhang2015deep}
Sixin Zhang, Anna~E Choromanska, and Yann LeCun.
\newblock Deep learning with elastic averaging {SGD}.
\newblock In {\em Advances in Neural Information Processing Systems}, pages
  685--693, 2015.

\end{thebibliography}
\section{Appendix: Detailed Derivations}

\subsection{Detailed Derivation of the Forward Pass}

\label{forwardpass}
During the forward pass for data parallel, each process reads a mini-batch size of $B/P$ input images and the calculations
are performed as follows:

\begin{equation*}
  Y_i = WX_i,
\end{equation*}
where $X_i$ and $Y_i$ is the $i$th column of X and Y, respectively. Here, $W$ is shared and no communication is needed. However,
in the model parallel case we have:

\begin{equation*}
  Y_{p,i} = W_pX_i,
\end{equation*}
where $p$ denotes the process id, $W_p$ is the fraction of weights in each process, and $Y_{p,i}$
is the fraction of output activation computed locally.
This local component needs to be communicated via an all-gather operation to concatenate all partial activations for the
next layer's computation.


\subsection{Detailed Derivation of Backpropagation}
\label{backprop}
During backpropagation, the gradient of the loss functional with respect to output activations $Y$ is given ($\Delta_Y = \frac{\d \mathcal{J}
}{\d Y}$), and one has to compute
the gradient with respect to the weights ($\Delta_W)$ as well as the input feature map ($\Delta_X)$. The latter is needed for propagating the gradient
to lower layers. We use capital letters for input and output activation as we are mostly interested in the mini-batch setting $B>1$. Using chain rule we have:

\begin{equation*}
   \Delta_W = \frac{\d\J}{\d W} = \sum_{i=1}^{B}\frac{\d\J}{\d Y_i}\frac{\d Y_i}{\d W} = \sum_{i=1}^{B}\frac{\d\J}{\d Y_i} X_i^T = \Delta_Y X^T,
\end{equation*}

Now in the distributed case, we have:

\begin{align*}
  \frac{\d\J}{\d W}^p = \sum_{i=1}^{B/P_c}\frac{\d\J}{\d Y_i} X_i^T,\\
  \frac{\d\J}{\d W} = \sum_{k=1}^{P_c}\frac{\d\J}{\d W}^p.
\end{align*}
where $p$ is the process id. Notice that the last step requires an all-reduce between $P_c$ processes, but no communication is needed for the model
parallel part as the input activation is already communicated via the all-gather collective of forward pass.
To backpropagate the gradient, one needs to compute $\Delta_X = \frac{\d\J}{\d X}$ as well. We derive it for one column of $\Delta_X$ below as each column can be computed independently:

\begin{equation*}
  \Delta_{X_i} = \frac{\d\J}{\d X_i} = \frac{\d\J}{\d Y_i}\frac{\d Y_i}{\d X_i} = W^T\frac{\d\J}{\d Y_i} = W^T \Delta_{Y_i},
\end{equation*}
\noindent Here in the distributed model parallel part, the weight matrix is distributed among $P_r$ processes.
To backpropagate the gradient, every process computes its contribution to the gradient followed by an all-reduce collective between $P_r$ processes:

\begin{align*}
  \frac{\d\J}{\d X_i}^p = W^T\frac{\d\J}{\d Y_i},\\
  \frac{\d\J}{\d X_i} = \sum_{k=1}^{P_r}\frac{\d\J}{\d X_i}^k. 
\end{align*}

\end{document}